\documentclass[letterpaper, 10 pt, conference]{ieeeconf}  

\IEEEoverridecommandlockouts                              
\usepackage{amssymb}
\usepackage{graphicx,amsmath}
\usepackage{lineno}
\usepackage{caption,url}
\usepackage{float}
\usepackage{epstopdf}

\usepackage{subcaption}
\usepackage{gensymb}
\usepackage[colorlinks]{hyperref}
\hypersetup{
    colorlinks=true,
    linkcolor=blue,
    filecolor=magenta,      
    urlcolor=cyan,
}

\title{\LARGE \bf
Learning Stable Manoeuvres in Quadruped Robots from \\  Expert Demonstrations
}

\author{Sashank Tirumala, Sagar Gubbi, Kartik Paigwar, Aditya Sagi, \\ Ashish Joglekar,  Shalabh Bhatnagar, Ashitava Ghosal, Bharadwaj Amrutur, Shishir Kolathaya 
\thanks{*This work has been funded by Robert Bosch Center for Cyber-Physical Systems (RBCCPS), Indian Institute of Science (IISc), Bengaluru.}
\thanks{S. Tirumala is with the Department of Engineering Design at the Indian Institute of Technology-Madras, Chennai {\tt\small email: {stsashank6} at gmail.com}} \\
\thanks{S. Gubbi, B. Amrutur are with the department of Electrical Communication Engineering and RBCCPS, K. Paigwar, A. Sagi, A. Joglekar are with RBCCPS, A. Ghosal is with the department of Mechanical Engineering and RBCCPS,  and S. Bhatnagar, S. Kolathaya are with the department of Computer Science \& Automation and RBCCPS, IISc, Bengaluru, India
 {\tt\small email: \{sagar, kartikp, adityavarma, ashishj, shalabh, asitava, amrutur, shishirk\} at iisc.ac.in}
 }
}%

\begin{document}

\maketitle
\thispagestyle{empty}
\pagestyle{empty}

\begin{abstract}
With the research into development of quadruped robots picking up pace, learning based techniques are being explored for developing locomotion controllers for such robots. 
A key problem is to generate leg trajectories for continuously varying target linear and angular velocities, in a stable manner.
In this paper, we propose a two pronged approach to address this problem. First, multiple simpler policies are trained to generate trajectories for  a discrete set of target velocities and turning radius. These policies are then augmented using a higher level neural network for handling the transition between the learned trajectories. Specifically, we develop a neural network based filter that takes in target velocity, radius and transforms them into new commands that enable smooth transitions to the new trajectory. This transformation is achieved by learning from expert demonstrations. An application of this is the transformation of a novice user's input into an expert user's input, thereby ensuring stable manoeuvres regardless of the user's experience.  
Training our proposed architecture requires
much less expert demonstrations compared to standard neural network architectures. Finally, we demonstrate experimentally these results in the in-house quadruped Stoch 2. 
\end{abstract}

\textbf{Keywords:} \textit{Quadrupedal walking, Reinforcement Learning, Random Search, Gait transitions}

\section{Introduction}

The domain of quadrupedal research has reached industrial markets today with quite a few research labs/companies successfully commercializing their quadruped robots \cite{Anymal2016Marco}, \cite{dhaivatdesigndevelopment}, \cite{grimminger2019open}.
Similar to driving a car, controlling a quadruped robot has a steep learning curve that a novice user must struggle through. Assuming you are given an interface to control the velocity of the center of mass of a quadruped robot, rapid changes in said velocity will cause the robot to topple. This instability is not safe and could permanently damage the robot. However an expert user will be capable of performing the desired rapid velocity changes through experience that he gained through practice. Is it possible then to augment the novice user's commands such that it represents the expert user's manoeuvres?   

\begin{figure}[t!]
\centering
\vspace{2mm}
\includegraphics[width = 0.8\linewidth] {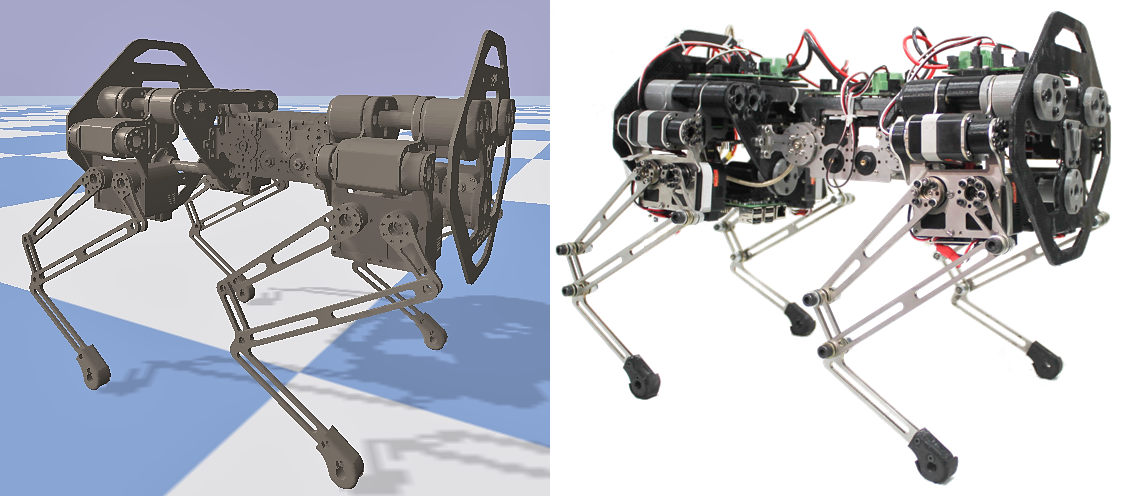}
\caption{Figure showing the custom built quadruped robot, Stoch 2. Simulated version is shown on the left, and the actual hardware is shown on the right.}
\label{fig:PyBullet}
\end{figure}

Quadruped Locomotion is a challenging research problem that has been studied for decades starting from the GE walking truck in 1960's \cite{getruck} . A slew of techniques have been used---inverted pendulum model based controllers \cite{raibert1986legged}, zero-moment point based controllers \cite{WinklerFast2017}, hybrid zero dynamics \cite{hzd_grizzle}, model predictive controllers \cite{Kim2019HighlyDQ}, deep reinforcement learning \cite{Hwangboeaau5872}---to name a few. These techniques provide an interface to control the center of mass velocity over a rough terrain \cite{Kim2019HighlyDQ}, \cite{Hwangboeaau5872}. Relatively little work has been done on handling rapid changes in the desired center of mass velocity. We propose to tackle this as a behaviour cloning problem. In particular, we have access to expert input that is capable of performing rapid changes in a stable and safe manner. We also have access to a novice user's input that is unsafe for the robot. We aim to train a neural network so that it takes a novice user's input and transforms it to an expert user's input. 
We validate our neural network by demonstrating rapid changes in linear ($0.0$ $m/s$ to $0.6$ $m/s$) and angular velocity ($\frac{-2\pi}{3}$ $rad/sec$ to $\frac{2\pi}{3}$   $rad/sec$) in our in-house quadruped robot Stoch $2$. 
\subsection{Related Work}
An omni-directional quadruped robot controller requires two parts: stable leg trajectories for different motions such as walking forward, turning etc., and techniques to transition between these trajectories when required. In the literature there has been more focus on the first problem of generating stable leg trajectories. \cite{kohl2004policy} first used policy gradient algorithms to learn optimal end-foot trajectories for the Sony AIBO-Quadruped Robot. 
\cite{Degrave2013Comparing} used particle swarm optimisation to decide the parameters for a few different trajectories including turning on the Oncilla Quadruped Robot. Our current work borrows from the turning strategies described in the above paper. \cite{Jie2018Sim} first demonstrated techniques to learn a quadruped controller in simulation and deploy it on the actual robot. \cite{Trajectory2019Kolathaya} modified the reinforcement learning algorithm to learn the parameterized leg trajectories quickly while \cite{tirumala2019gait} constrained the network architecture to speed up learning. In this paper we combine \cite{Trajectory2019Kolathaya} with a sample efficient learning algorithm, augmented random search  \cite{mania2018simple}, to learn a hundred trajectories in simulation and deploy the learnt trajectories on our robot.

Transitioning between stable trajectories for quadrupeds was first studied by \cite{Bruce2002Fast}, in which analytical equations were used to stitch Hermite splines. \cite{Ma2005Omnidirectional} developed a similar analytical framework to transition between a number of different trajectories for slow static walking. These methods, despite being effective, were limiting the number and speed of allowable transitions. 
The MIT Cheetah \cite{Bledt2018MIT} proposed a PD controller to track varying center of mass velocities combined with Raibert's controller \cite{raibert1986legged} to ensure stability of the robot. In general such controllers require expensive series elastic actuators or direct-drives with torque control.  
\cite{PMTG} trained a reinforcement learning agent capable of tracking a specific desired velocity profile while \cite{Hwangboeaau5872} trained a reinforcement learning controller capable of tracking any combination of linear and angular velocities. Our work is more adjacent to the above works, where we would like to replace unstable commands given by a novice user with stable commands given by an expert user. Our controller can sit on top of the above proposed controllers and does not have any specific actuator requirements. 
This controller will be based on a neural network based filter, which has the ability to generalize well. We will also show that the types of transitions executed are nonlinear, and linear filters do not yield the same result (see \ref{ssec:filter} ahead).
\subsection{Contributions}
In this paper we extend the work on trajectory generation by using the analytical equations in \cite{Degrave2013Comparing}, \cite{Trajectory2019Kolathaya}, to constrain the action space of our learning agent, thereby learning $100$ trajectories in $2$ hours on an Intel Core i7 processor. Our learnt trajectories demonstrate omni-directional motion in our robot Stoch $2$. To transition between different trajectories, we exploit the knowledge that a human expert has in tele-operating the quadruped robot. We collect expert demonstrations and train a neural network to convert novice user input to expert user input. Our neural network has a novel architecture that consists of two non-trainable layers. It also consists of a classification network that takes in the user input and outputs the probability of choosing a particular filter at every time-step. This probability is used to calculate a weighted average of the filters, which is the final output of the network. By restricting our network to only output weighted averages of different filters at every time-step, we require much less training data to generalize, compared to fully connected dense neural networks and convolutional neural networks. This solves the bottleneck of collecting expert data that is prohibitive on real robot systems. Our final network is capable of generalizing with five expert demonstrations. We validate our network by testing it on our quadruped robot. 
\section{Robot model and control}
\label{sec:background}

\begin{figure}
    \centering
    \includegraphics[width =0.8 \linewidth]{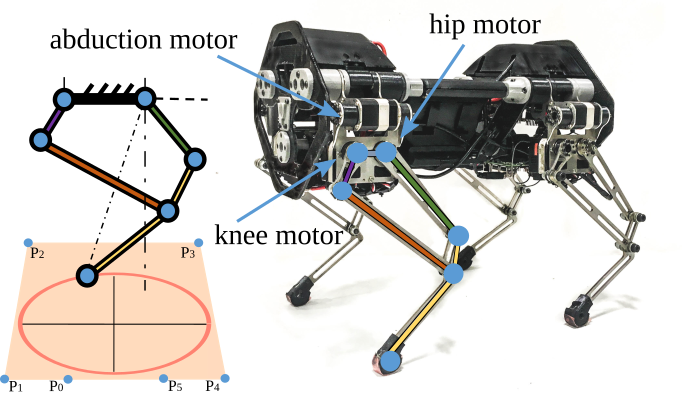}
    \caption{A five-bar mechanism is used as the leg. This mechanism is actuated by the motors located at the main torso of the robot. The trajectory followed by the foot is parameterized by a rational Bezi\'er curve with control points $P_0,...,P_5$ as shown. }
    \label{fig:IK_S2}
\end{figure}

In this section, we will discuss 
our custom built quadruped robot Stoch 2. Specifically, we will provide details about the hardware, the associated model, and the trajectory tracking control framework used for the various gaits of the robot.


\subsection{Kinematic Description}\label{subsec:kinematic}

\textit{Stoch $2$}  is  a  quadruped  robot designed  and developed in-house at the Indian Institute of Science (IISc), Bengaluru, India. 
In this robot, each leg comprises of a parallel five-bar linkage mechanism where two of the links are actuated and the end effector (i.e., the foot) is capable of moving safely (without encountering singular configurations) in a trapezoidal work-space as shown in Fig. \ref{fig:IK_S2}.
The calculation of the leg work-space and it's inverse kinematic details can be found in \cite{dhaivatdesigndevelopment}. 
In this paper, we focus on realizing trajectories of the feet in Cartesian coordinates on a fixed plane. The plane is chosen such that it passes through the center of the hip joint and is constrained by the three dimensional trapezoidal prism work-space of each foot. 


\subsection{Control Framework}
Our control framework takes in the turning radius and the heading velocity
as input (mapped to joystick analog values), and then outputs the end-effector (end-foot) trajectory. This end-foot trajectory is sent to our inverse kinematics engine, that calculates the desired motor commands for each of the $12$ servo motors in our robot. 
Stable end-foot trajectories are found in simulation.
We chose to parameterize the end-foot trajectory of each foot by a rational-Bezi\'er curve.
Rational Bezi\'er Curves have attractive properties over other alternatives like cubic splines as they do not have self-intersecting curves, and are guaranteed to lie within the convex hull of the control points. This ensures that the trajectories are always in the work-space of the end-effector. The control points of the rational Bezi\'er in 3D space are chosen analytically and lie on the boundaries of the robot leg-workspace as shown in Fig. \ref{fig:IK_S2}. The weights of each of the control points are the parameters that we aim to find through our learning framework in simulation. 
We chose a $6^{th}$ order Bezi\'er curve for the swing phase of our leg, and a $2^{nd}$ order Bezi\'er curve (straight line) for the stance phase motion of our leg. This choice was made to represent optimal half-boat shaped trajectories for mechanical walkers as described in  \cite{shigley1960mechanics}.
Given $n+1$ control points, denoted by $P_i$, $i=0,1,\dots,n$, and the weights, denoted by $w_i$, we have the Bezi\'er curve given by
$$P(t) = \frac{\sum_{i=0}^{n} {n \choose i} t^i (1-t)^{n-i} P_i  w_i}{\sum_{i=0}^{n} {n \choose i} t^i (1-t)^{n-i} w_i},$$
where $t \in [0,1]$, and $w_i > 0 \quad \forall i$. In practise, computing $n \choose i$ is computationally expensive, so we use a recursive implementation of the De-Casteljau Algorithm. The analytical equations that determine the control points of our Bezi\'er curve is based upon a simplified model for our robot locomotion that we formulated through experimentation. Our experimentation showed that turning is more effective if abduction and step-size for each leg are dynamically varied. This result corroborates well with the analysis in \cite{Degrave2013Comparing}. In particular, each end-foot trajectory is restricted to a plane tilted by an angle $\phi$ about the $z$ axis in such a way that
$$
\phi_{i} = \tan^{-1}\left (\frac{\frac{L}{2}}{|r| + \delta_{i}*sign(r)* \frac{W}{2}}) \right),
$$
where $r$, the radius of curvature, is related to the desired linear velocity $v$ and angular velocity $\omega$ as $r = \frac{v}{\omega}$. $W$ ($0.12m$) stands for the width of the chassis of the robot and $L$ ($0.24m$) stands the length of the chassis of the robot.
 $sign$ stands for signum function, $\phi_{i}$ is the angle made by the $i^{th}$ leg with the vertical axis and  $\delta_{i}$ is a constant that takes the following values for the four legs: $$\delta_{fl} = 1,\hspace{0.1cm} \delta_{fr} = -1,\hspace{0.1cm}  \delta_{bl} = 1,\hspace{0.1cm} \delta_{br} = -1 , $$ where $fl,fr,bl$ and $br$ stand for the front-left, front-right, back-left and back-right legs respectively. Similarly, each leg must also have a step length $(sl)$ equal to
$$sl_{leg} = v * \left ( abs(r) + \delta_{leg}*\frac{W}{2}\right )/{abs(r)}, $$
where $v$ stands for the desired average speed of the robot for $1$ second. Note that $r$ can be both positive and negative (for left-right turn commands). Given the step-lengths and the plane angles of each leg, we can determine the control points of the Bezi\'er curve that each leg follows. As shown in Fig. \ref{fig:IK_S2} above we place six control points such that they lie on the boundaries of our trapezoidal work-space, and lie on the plane that makes an angle $\phi_{leg}$ about the z-axis. 
The first and last control points are chosen such that they are symmetric about the center of the work-space and the distance in-between them is equal to the step-length of the respective leg.   

\subsection{Trajectory Generation Framework}
The weights of the Bezi\'er curve decide the overall shape while the control points limit the search space of all possible shapes. We formulate this trajectory generation as an optimization problem where we aim to find the weights $w_i$ such that a certain cost function $J$ is minimized. For a particular linear and angular velocity our weights $w_i$ are held constant. We chose to define our cost function $J$ as
$$J = \sum_{t = 0} ^{T}{(E + p_{\phi} + p_{\psi})},$$
where $E$ is the energy consumed per step, $p_{\phi}, p_{\psi}$ are the penalties related to rolling and pitching of the robot, and $T$ is the number of control time-steps for a single simulation. $E$ is given by 
$$E = \sum_{i = 0}^{n} \omega_i \tau_i \triangledown t. $$ Here $\omega_i$ is angular velocity of $i^{th}$ motor, $\tau_i$ is torque of $i^{th}$ motor and $n$ is the total number of motors. These values are calculated by the simulation software during the training process. 
Here $p_{\phi} = 0.1*|\phi|, p_{\psi}=0.05*|\psi| $, where $\phi$ and $\psi$ are roll and pitch angles about the x-z axes of the center of mass respectively.
To optimize for $w_i$, many different algorithms can be used. We chose to use Augmented Random Search \cite{mania2018simple}, a well-known training algorithm for linear policies.

\section{Trajectory Transition Framework}
Having defined the model and the control methodology, we are now ready to discuss the trajectory transitioning framework of our robot. This part is divided into two sections where we describe the problem formulation, the neural network architecture and the training process.
\subsection{Problem Formulation}
We are interested in aggressive manoeuvres where a naive approach will cause the robot to lose balance and fall.
The user can input linear and angular velocity ($v, \omega$) through joystick commands. In our experimentation, it was easier to control the linear velocity and radius of curvature $r = \frac{v}{\omega}$. 
We normalized the joystick values to the range $(-1, 1)$. Positive $r$ indicates moving rightwards, negative $r$ indicates moving leftwards, while positive $v$ indicates moving forward and negative $v$ indicates moving backward. 

\begin{figure*}\centering
  \includegraphics[width=0.95\textwidth]{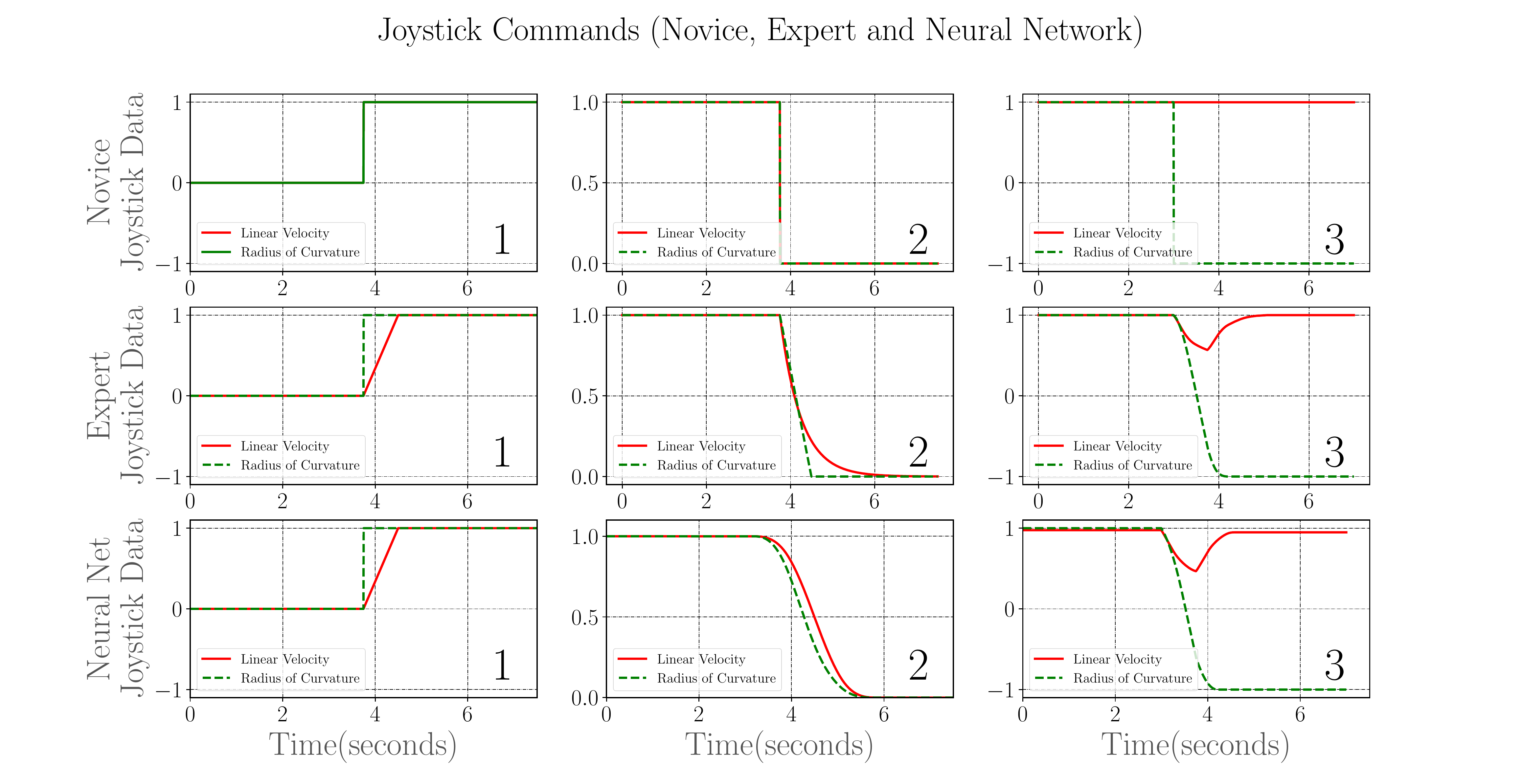}
  \caption{ Green dotted line stands for the radius of curvature while the red solid line is the heading velocity. Both values are normalized between $-1$ and $1$. Joystick data for novice is at the top, and expert data is at the middle for the three described manoeuvres. Neural network output after training is at the bottom.}
  \label{fig:data_fig}
\end{figure*}

To demonstrate our trajectory transition framework, we consider three complex manoeuvres that can potentially damage the robot. These manoeuvres are: $1$ - rapidly reaching maximum linear velocity and curvature radius, $2$ - abruptly coming to a halt from maximum linear velocity and curvature radius, $3$ - rapidly changing radius of curvature direction.
Then for each of the above manoeuvres we collected novice and expert joystick data as shown in Fig. \ref{fig:data_fig}. We observed that there exists a complex relationship between the radius and velocity inputs that one dimensional filters cannot reproduce. In particular, manoeuvre $3$ requires $v$ to drop whenever $r$ sharply changes, which cannot be replicated with a simple filter. A simple filter would cause $r$ to gradually reduce from $1$ to -$1$ and not affect $v$. In manoeuvre $1$ and $2$ both $v$ and $r$ change at different rates. In manoeuvre $1$, $r$ changes instantly while v moves gradually, while in $2$, $r$ reduces gradually while $v$ reduces exponentially. Without knowledge of the expert trajectories, choosing an appropriate filter for each manoeuvre is not straightforward. In an analytical approach, as the number of manoeuvres increase, more effort is required to design filters, while some manoeuvres like $3$ cannot be recreated with linear filters. A neural network bypasses these difficulties and can scale to as many expert demonstrations as required with no additional effort. Hence, our goal now is to train a neural network that is capable of taking the novice joystick-data, as shown in Fig. \ref{fig:data_fig}, and converting it to the expert joystick-data.

\subsection{Neural Network Architecture and Training}
\begin{figure}
\centering\includegraphics[width = 0.45\textwidth]{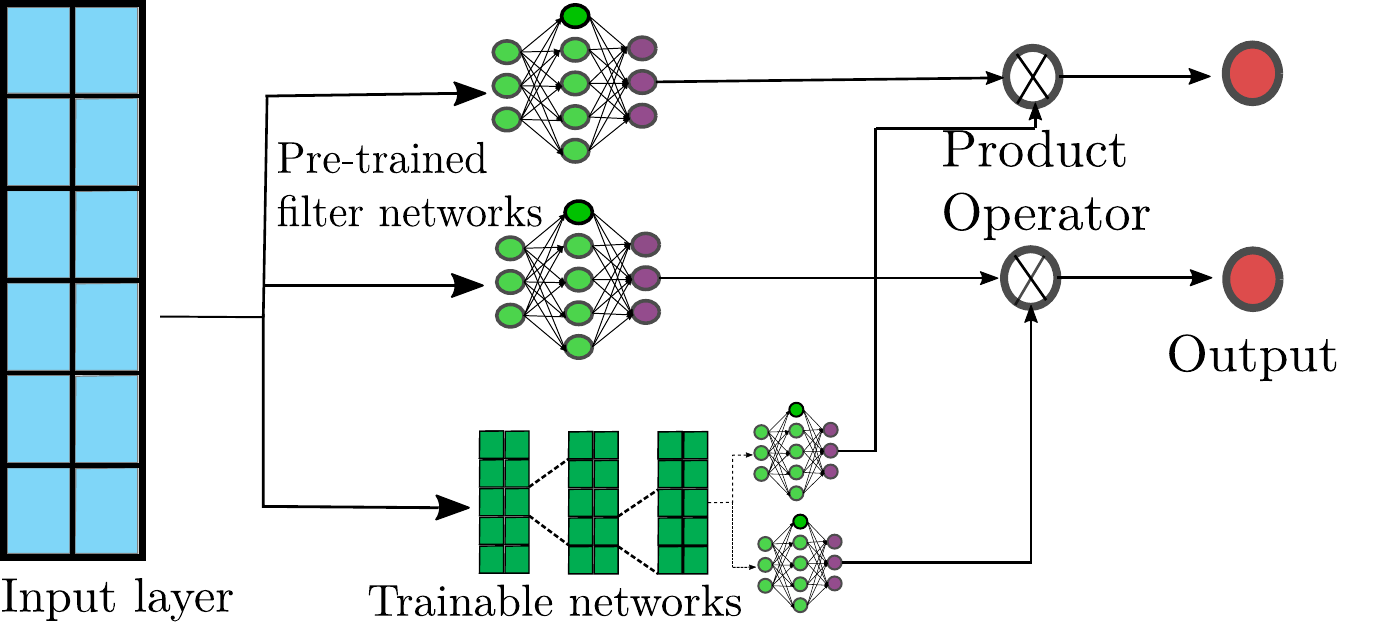}
  \caption{Visualization of the neural network. It splits into two parts, the top layer being non trainable that acts like the low pass filters and the bottom layer being trainable. The bottom layer is a combination of a deep convolutional net (10 layers) followed by 2 fully connected layers for each column of the input. The final output of the bottom layer is multiplied by the final output of the top layer (filter bank) to give the output of the neural network. The product operator is a dot product between two vectors of size $\mathbb{R}^{64}$.}
  \label{fig:nn_fig}
\end{figure}
The input to our neural network is the past $250$ joystick values from each analog stick and thus $x \in R^{250 \times 2}$. These values are from a duration of $1.25$ seconds. The output of our neural network is the linear velocity $v$ and radius of curvature $r$ for the current time-step. The first two layers of our neural network consist of non trainable layers that we call ``filter-banks". The weights of these layers are such that they act as simple filters for the joystick inputs. In particular the output of our non-trainable layers is a set of $64$ values that is a combination of low pass filtered versions of the input and scaled versions of the input. The low pass filters are moving average filters of window size $1,2,3,4,5,6,7,8,10,20,30,\dots,240$ and scaling filters multiply the input by a constant $k$ which varies from $0.9,0.9^{2},\dots,0.9^{33}$. 

In parallel, we have a 1D convolutional neural network (CNN) that takes  $x \in R^{250 \times 2}$  as input and outputs a value $y \in R^{128}$. Our CNN is $10$ layers deep with $6$ convolutional layers and $4$ max pooling layers with Re-Lu activation. The output of CNN is sent to two sets of fully connected dense layers with softmax activation function that outputs $64$ values that correspond to the probability of choosing a particular filter from the filter bank. The output of these dense layers is used to compute the weighted average of the filter banks which leads to the final output of the network. This is similar to the attention mechanism of neural networks. The visualization of this network is shown in Fig. \ref{fig:nn_fig} and we have released the code of this network implemented in tensorflow (provided in \ref{sec:conc}). We collect a single demonstration of novice and expert input for each desired manoeuvre (three shown in Fig. \ref{fig:data_fig}), and train the neural network upon these demonstrations. The learnt network is capable of generalising unseen data of similar shape but of different amplitudes and frequency unlike a fully connected neural network. The hyper-parameters used in our training process are shown in Table 1. The training was stopped once validation loss fell below $10^{-2}$.

\begin{table}[]
\centering
\begin {tabular}{|l|l|}
\hline
                    & Learning Rate              \\ \hline
Learning Rate       & 0.0002                     \\ \hline
L2-Regularization   & 0.000001                   \\ \hline
Dropouts \%         & 50                         \\ \hline
Activation Function & ReLu + Softmax(last layer) \\ \hline
Batch-size          & 64                         \\ \hline
Optimizer           & Adam                       \\ \hline
\end{tabular}
\caption{Hyperparameters of neural network}

\end{table}

\section{Experimental Results}
In this section we provide results to show: improvement of our trajectory generation framework to existing techniques, comparison with standard non-neural network based filters and the improvements of our proposed neural network architecture to standard neural network architectures.
\label{sec:experiment}
\subsection{Trajectory Generation Framework Experiment}
\begin{figure}\centering
\includegraphics[width=0.5\textwidth]{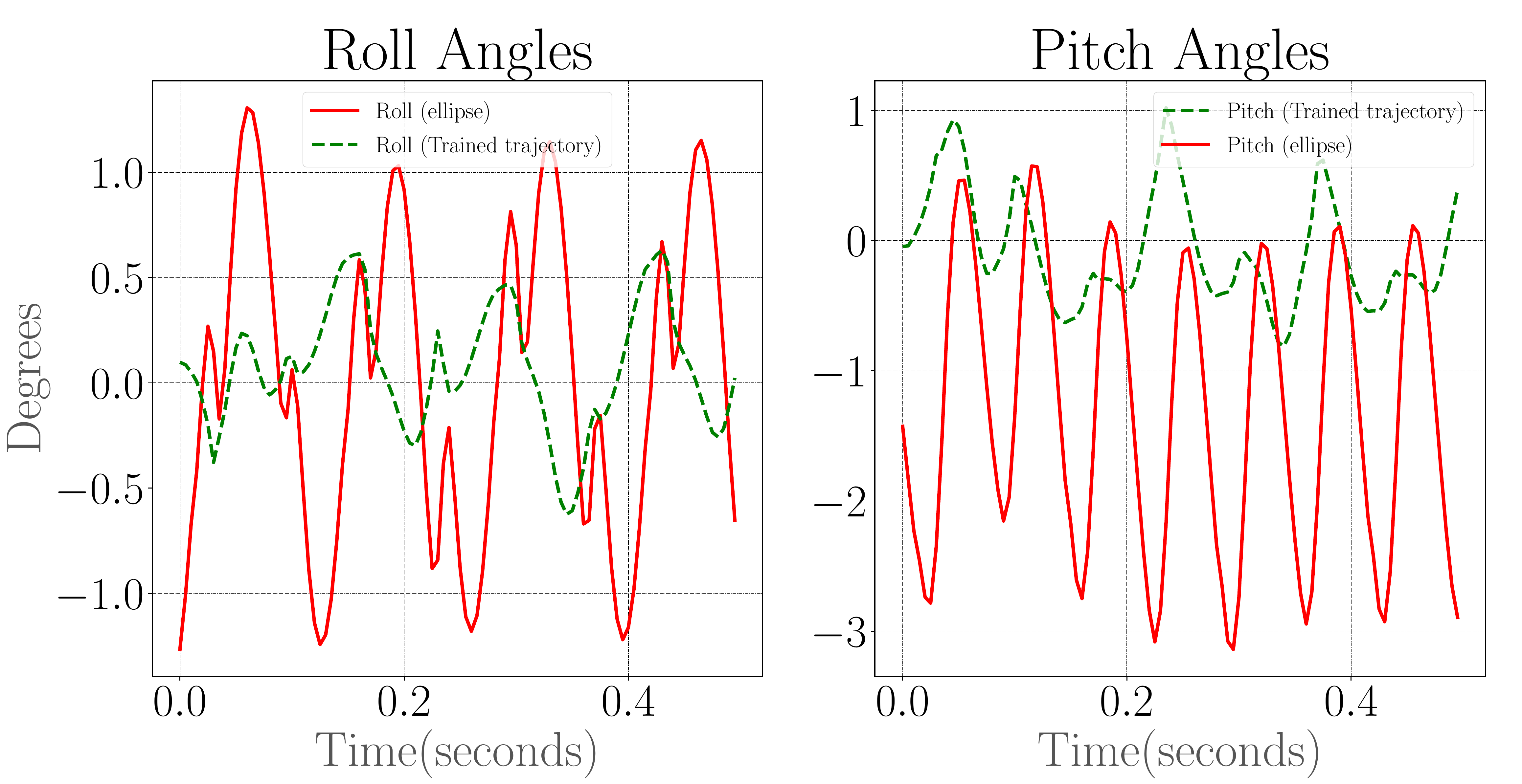}
  \caption{Variation of roll (left) and pitch (right) angles versus time in the robot. Green is the learnt trajectory while red is the default elliptical trajectory.}
  \label{fig:roll_data_fig}
\end{figure}

The main goal of our trajectory generation framework is to find stable trajectories for our robot quickly. Stability of the trajectory is measured using the pitch and roll angles of our robot's body in simulation. A more stable trajectory will have lesser amplitude of oscillation. Compared with a default elliptic trajectory our learnt agent had approximately $50\%$ and $66\%$ lower amplitude of oscillation of roll and pitch angles respectively as seen in Fig. \ref{fig:roll_data_fig}. This was expected as the pitch and roll penalties were explicitly added in our cost function during the training process. Generating a single trajectory takes a few minutes on an Intel i7 core PC, and generating the entire set of trajectories for all velocities and radius of curvature took about $2$ hours. 
\subsection{Comparison with Standard Filters} \label{ssec:filter}
\begin{figure}\centering
\includegraphics[width=0.45\textwidth]{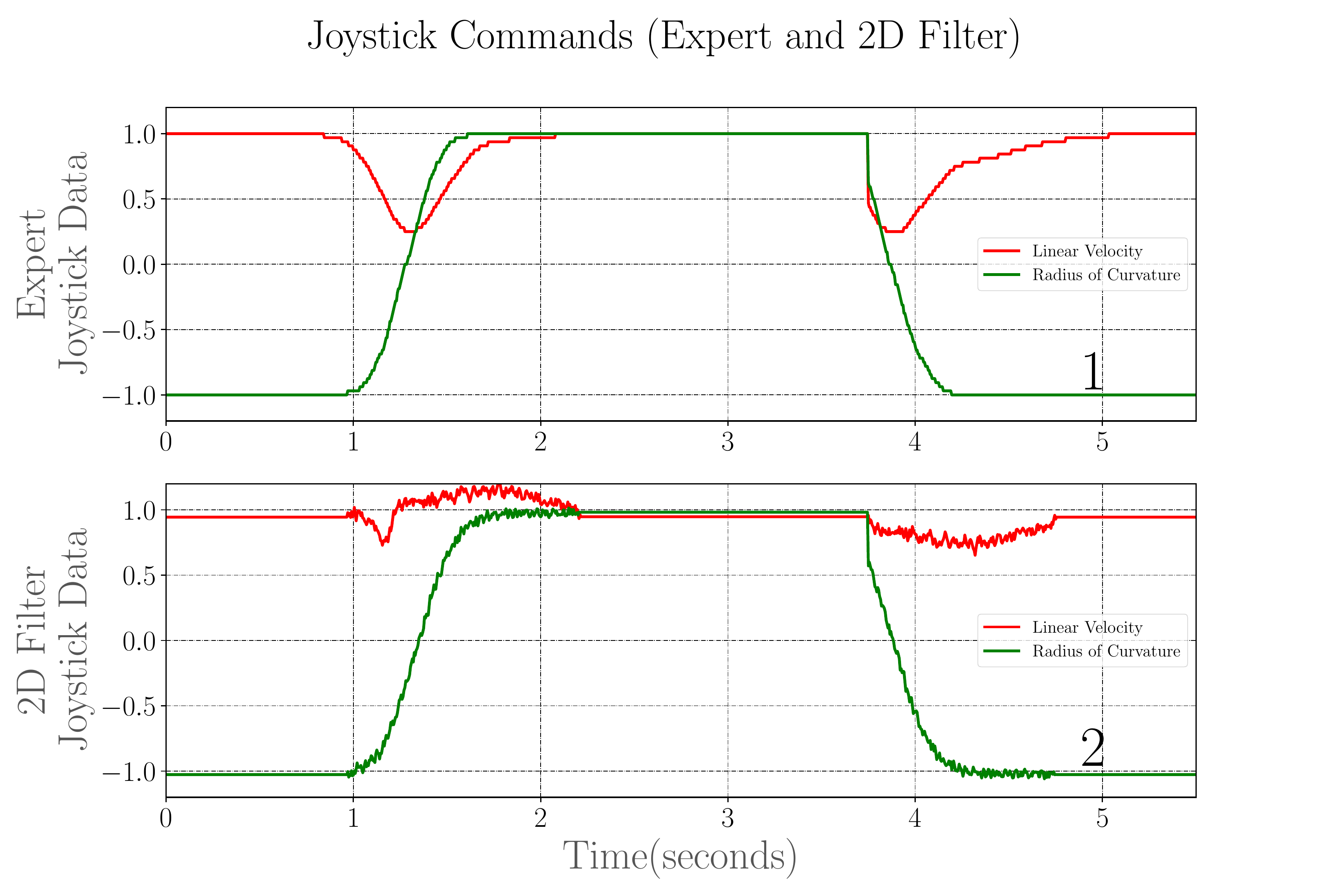}
  \caption{Comparison between actual expert data and a standard 2D convolutional filter. As we can see a standard 2D filter struggles to copy the expert and exhibits properties of linearity that are undesirable}
  \label{fig:filter_data_fig}
\end{figure}

Standard convolutional 2D filters suffer from a number of problems that make them unsuitable to the current application. They are unable to copy complex transitions like transition 3 as shown in Fig. \ref{fig:filter_data_fig}. Often the output of such a filter is a crude approximation of the actual expert output. Further, these filters have linear properties such as output superposition and scaling, which our expert output does not follow. To illustrate, consider the change of radius from $-1$ to $1$ and $1$ to $-1$ as shown in Fig. \ref{fig:filter_data_fig}. Both of these transitions require the same dip in velocity as the transitions are symmetric (left turn $~$ right turn). However a linear filter trained on transition $3$ will cause a peak in velocity for the opposite direction, which actively destabilizes the system. Similarly at lower velocities, a linear filter will output a scaled version of the strategy used at higher velocities while in reality an expert user will often pursue a very different strategy. Thus nonlinear alternatives, specifically neural network based alternatives are pursued. 
\subsection{Neural Network Architecture}
A standard issue with techniques that imitate experts is that the collection of expert data tends to be time consuming and costly. Thus we designed a network that requires as little expert demonstrations as possible.
Here we aim to measure the generalizability of our neural network compared to standard neural network architectures. Generalizability is a broad term and since we are doing only supervised learning we cannot expect our network to truly imitate an expert in completely unforeseen situations. Instead by generalizability we mean two major properties of our expert data: first, our data is a time invariant system. By this we mean that if a novice input is delayed by $t$ seconds then our expert output should also be delayed by $t$ seconds. No other changes to the output is necessary.  The second is our data is approximately scale invariant. This means that if we multiply the novice input by a constant factor $k$, then our expert output is also approximately multiplied by a constant factor $k$. This need not be strictly followed as experts tend to follow different strategies at different speeds, but broadly our network should be capable of handling scale invariance within a limit. 

\begin{figure}[ht]
\centering
\includegraphics[width=0.49\textwidth]{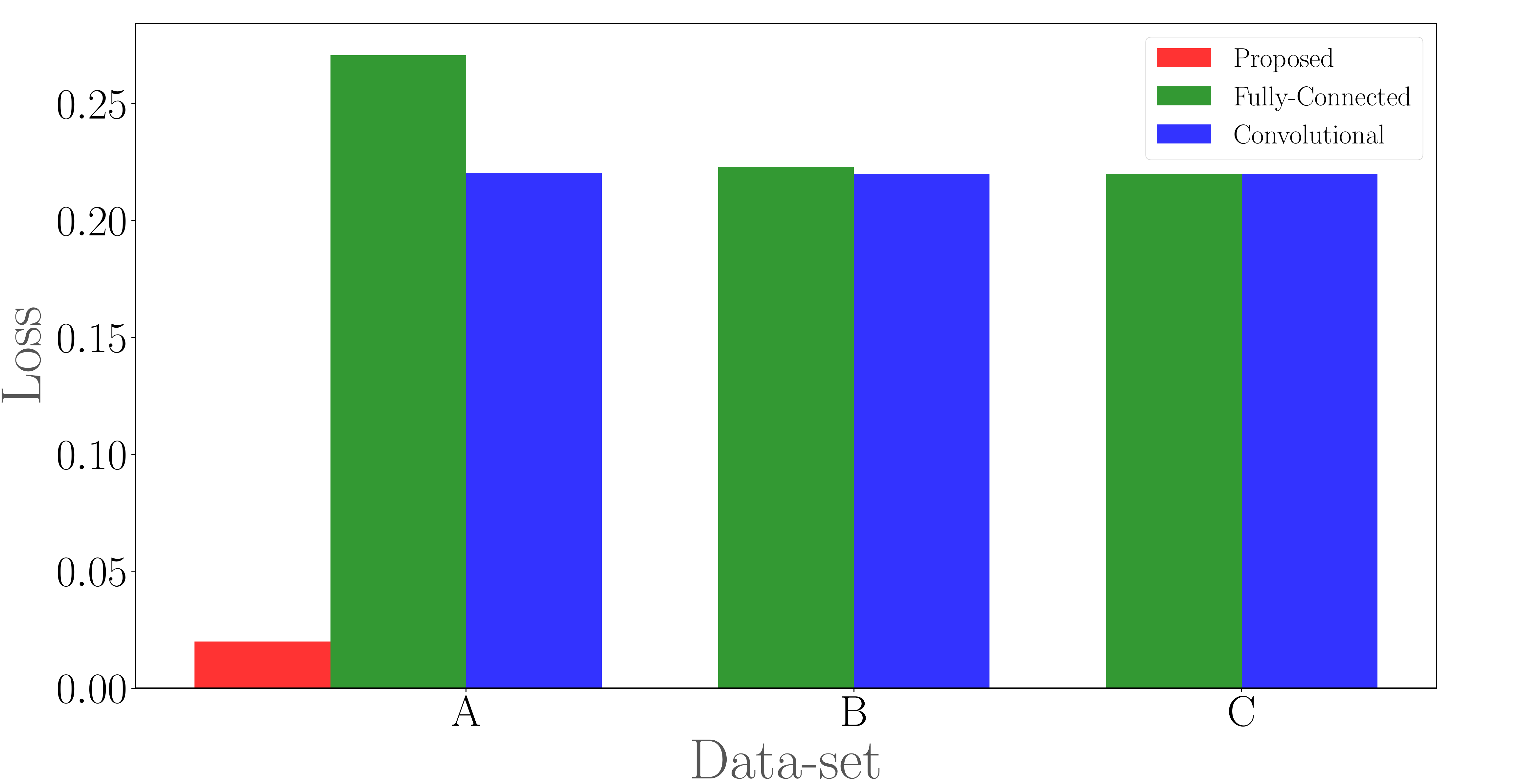}
  \caption{Variation of validation-loss with training data. All networks reach very low training error however our network generalizes better to the validation data. Dataset A has $5$ demonstrations for training, while B, C have $10$ and $20$ demonstrations respectively. Note that losses for B, C with the proposed network are very small to be noticed.}\vspace{-2mm}
  \label{fig:tr_data_fig}
\end{figure}

To measure the capabilities of our network, we measure the validation loss of our network compared to standard neural network architectures.
We do so by first collecting a number of expert trajectories. Then we augment the data by temporally delaying it, and scaling it to about $100$ different trajectories. We compare our proposed architecture with two common architectures: a) A neural network with $10$ fully connected dense layers, and b) a neural network with $10$ fully connected convolutional layers. We use $5$, $10$, and $20$ demonstrations respectively as our training data-set and measure the loss over the entire training and validation data-set of $100$ demonstrations. The results are shown in the Fig. \ref{fig:tr_data_fig}. As we can see from the above results, the validation loss of a fully connected or a CNN barely decreases with increase in size of the data-set. This shows that these architectures cannot generalize well. However our proposed neural network architecture has $2 \times 10^{-4}$ validation loss with $10$ demonstrations, making it more sample efficient than standard neural networks. 

%

\section{Conclusion and Future Works} \label{sec:conc}
We have presented a trajectory generation and transition framework that is easy to use and applicable for omni-directional motion of quadrupedal robots. Trajectory transitions are learnt from demonstrations of an expert user. Trajectory transitions are achieved by a neural network that uses a unique architecture, which is much more suited to mimicking filters than fully connected or convolutional neural networks. As an application, we show how a novice user's joystick command is converted to safe commands.
In future, we would like to generate trajectories for more complex terrains such as stair climbing and uphill slopes. Automating the generation of trajectories given a robot model to quickly generate controllers for different quadruped robots is also an exciting research direction. The video submission accompanying this paper is shown here: \href{https://youtu.be/LRbHetp0dcg}{https://youtu.be/LRbHetp0dcg}, and the code is released here: \href{https://github.com/sashank-tirumala/stoch_robot_test2}{https://github.com/sashank-tirumala/stoch\_robot\_test2}.







\bibliographystyle{IEEEtran}
\bibliography{references}

\end{document}